\documentclass[hyperref={colorlinks,linkcolor=blue,anchorcolor=blue,citecolor=blue}]{IEEEtaes}

\usepackage{cite}
\usepackage{orcidlink}
\usepackage{color,array,amsthm}
\usepackage{graphicx}
\usepackage{amsmath,amsfonts}
\usepackage{algorithmic}
\usepackage{algorithm}
\usepackage{booktabs}
\usepackage{makecell}
\pdfoutput=1

\begin{document}
\definecolor{Orange}{RGB}{255,69,0}

\title{Progressive Multi-Level Alignments for Semi-Supervised Domain Adaptation SAR Target Recognition Using Simulated Data}

\author{XINZHENG ZHANG~\orcidlink{0000-0003-0170-992X}}
\member{Member, IEEE}
\affil{Chongqing University, Chongqing, China}

\author{HUI ZHU~\orcidlink{0009-0007-7410-9884}}
\affil{Chongqing University, Chongqing, China}

\author{HONGQIAN ZHUANG~\orcidlink{0009-0007-4900-7843}}
\affil{Chongqing University, Chongqing, China}

\receiveddate{Manuscript received August 16, 2024; revised August 16, 2024; accepted August 16, 2024. This work was partially funded by the National Science Foundation of China (Grant 61301224) and partly by the National Science Foundation of Chongqing (Grant cstc2021jcyj-msxmX0174). \textbf\emph{(Corresponding author: Xinzheng Zhang.)}\\
Xinzheng Zhang, Hui Zhu and Hongqian Zhuang are with the School of Microelectronics and Communication Engineering, Chongqing University, Chongqing 400044, China (e-mail: zhangxinzheng@cqu.edu.cn; Victory-ZH@outlook.com;zhuanghongqian@stu.cqu.edu.cn).\\
Xinzheng Zhang is also with the Chongqing Key Laboratory of Space Information Network and Intelligent Information Fusion, Chongqing 400044, China.}

\maketitle

\begin{abstract}
Recently, an intriguing research trend for automatic target recognition (ATR) from synthetic aperture radar (SAR) imagery has arisen: using simulated data to train ATR models is a feasible solution to the issue of inadequate measured data. To close the domain gap that exists between the real and simulated data, the unsupervised domain adaptation (UDA) techniques are frequently exploited to construct ATR models. However, for UDA, the target domain lacks labeled data to direct the model training, posing a great challenge to ATR performance. To address the above problem, a semi-supervised domain adaptation (SSDA) framework has been proposed adopting progressive multi-level alignments for simulated data-aided SAR ATR. First, a progressive wavelet transform data augmentation (PWTDA) is presented by analyzing the discrepancies of wavelet decomposition sub-bands of two domain images, obtaining the domain-level alignment. Specifically, the domain gap is narrowed by mixing the wavelet transform high-frequency sub-band components. Second, we develop an asymptotic instance-prototype alignment (AIPA) strategy to push the source domain instances close to the corresponding target prototypes, aiming to achieve category-level alignment. Moreover, the consistency alignment is implemented by excavating the strong-weak augmentation consistency of both individual samples and the multi-sample relationship, enhancing the generalization capability of the model. Extensive experiments on the Synthetic and Measured Paired Labeled Experiment (SAMPLE) dataset, indicate that our approach obtains recognition accuracies of 99.63\% and 98.91\% in two common experimental settings with only one labeled sample per class of the target domain, outperforming the most advanced SSDA techniques.
\end{abstract}

\begin{IEEEkeywords}
SAR, target recognition, simulated data, domain adaptation.
\end{IEEEkeywords}

\section{Introduction}
Compared with optical imaging systems, synthetic aperture radar (SAR) can capture images with high resolutions in any weather condition. As a result, SAR has been applied extensively in both the military and private sectors. And, it has demonstrated great potential in the fields of terrain survey, environmental monitoring, and battlefield reconnaissance. Automatic target recognition (ATR) occupies a crucial position in SAR image processing and application, being significant for the fields of military intelligence.

The emergence of deep learning (DL) has brought remarkable progress in SAR ATR \cite{r1,r2,r3}, since DL can effectively extract high-level semantic features in contrast to traditional hand-craft features. There are several DL-based SAR ATR methods \cite{r3,r4,r5,r6,r7,r8,r9,r10,r11,r45,r46}, all of which achieve outstanding recognition performance. However, a successful DL model needs a vast amount of labeled samples for training. Nevertheless, since SAR targets are usually noncooperative, it is challenging to gather sufficient labeled data. Furthermore, the acquisition of experimental target data by real flight experiments is usually characterized by high economic cost and time-consuming, which further increases the difficulty of obtaining labeled data \cite{r12,r13}. Therefore, how to tackle the problem of limited training samples faced by SAR ATR is one of the main challenges in current research.

\subsection{Background}
In recent years, a compelling research trend for SAR ATR has emerged, that is, the utilization of simulated SAR data as a substitute for training DL models is a solution to the problem of insufficient measured data. However, despite the theoretical attractiveness of this solution, modeling inaccuracies in simulation generally lead to a significant domain gap between the simulated and measured data \cite{r14,r15}. The gap may lead to severe performance degradation in real-world applications, making it difficult to apply DL-based ATR models trained on simulated data to measured data. Consequently, there is an urgent need for the solution to bridge the domain gap between the simulated and measured data. To address this problem, a number of researchers have looked into techniques based on unsupervised domain adaptation (UDA) \cite{r15,r16,r28,r29,r47}. In the absence of labeled data on the target domain, UDA aims to efficiently align distributions between the simulation and measurement domains. By doing this, a more accurate and dependable ATR model may be trained merely using the simulated data. For example, Shi \textit{et al}. \cite{r15} proposed a UDA method for simulated data-aided SAR ATR, by carrying out alignments across domain-level and category-level granularities. 

In UDA, no data are labeled within the target domain, posing a great challenge to ATR models. In contrast, semi-supervised domain adaptation (SSDA) can excavate a minority of labeled samples in the target domain to guide cross-domain alignment, being superior than UDA \cite{r18}. For instance,  natural optical image classification under SSDA learning fashion can gain great accuracies with only one labeled sample per class in the target domain \cite{r17,r18}. It is noteworthy that the current SSDA methods are built on real optical datasets, whereas the properties of SAR images differ greatly from those of optical images. Thus, domain adaptation strategies specific to SAR images should be considered \cite{r15,r16,r19}. In particular, the domain shift is unique between the real and simulated SAR images. This is because the simulated SAR images are usually generated by physics-based electromagnetic (EM) modeling. The EM modeling is based on a fine CAD model of the observed target, followed by an image formation algorithm with the same radar parameters as the measured data. Hence, despite the domain shift existing between the real and simulated SAR data, there is abundant priori information in the former closely related to the latter. And, these two domains differ from general SAR image domains in terms of domain discrepancies. Therefore, it is essential to consider an SSDA method suitable for the simulated data-aided SAR ATR.

\subsection{Solution and Main Contributions}
To address the above challenges, we propose an SSDA approach based on progressive multi-level alignments. In the proposed framework, a novel data augmentation technique termed as the progressive wavelet transform data augmentation (PWTDA), is developed to carry out the domain-level alignment, based on deeply analyzing the image characteristic discrepancies between the simulation and measurement domains. In order to implement cross-domain category-level alignment, we design the asymptotic instance-prototype alignment (AIPA) strategy, by gradually facilitating the source domain samples close to the corresponding category prototypes of the target domain. Furthermore, the consistency alignment is introduced as another categorical alignment only in the target domain to increase the robustness and generalization of the entire model, according to the consistency regularization principle of semi-supervised learning. The proposed approach can effectively boost the accuracy of simulation data-assisted SAR ATR in the case of incredibly rare measured samples with labels.

The main contributions of the proposed SSDA approach for simulated data-assisted SAR ATR can be summarized as follows.

1) We present the PWTDA to achieve domain-level alignment, by blending the wavelet transform high-frequency sub-band components of images from the simulation and measurement domains. Specifically, the PWTDA is implemented in a progressive manner enhancing the diversity of the augmented samples and narrowing the domain gap gradually, which is conducted via adding reliable pseudo-labeled samples of the target domain batch-by-batch.

2) The AIPA is developed to force the samples of the simulation domain to approximate the associated category prototypes of the measurement domain, being beneficial for the category-level alignment. It is worth mentioned that the AIPA is also in a gradual manner, due to the fact that the class prototypes of the measurement domain are updated progressively. On the basis of the AIPA, we developed an asymptotic prototype alignment loss.

\subsection{Organization}
The following are the additional portions of this paper. The associated work is briefly described in Section \ref{sec2}. We describe the specifics of our suggested approach in Section \ref{sec3}. The experimental results and the related analysis are displayed in Section \ref{sec4}. The paper is discussed in Section \ref{sec5}. Finally, the paper is summarized in Section \ref{sec6}.

\section{Related Work}
\label{sec2}
\subsection{SAR ATR based on measured data using DL}
In contrast to traditional approaches, DL-based methods possess a significant feature extraction capability from sufficient measured training samples. Various DL models can achieve high accuracies and remarkable performance for SAR ATR.

In DL-based SAR ATR, supervised learning is a commonly used paradigm. Chen \textit{et al}. \cite{r1} pioneered a fully-connected network being trained on sufficient real SAR data for ATR. Without requiring a significant quantity of labeled training samples, Pei \textit{et al}. \cite{r48} demonstrated a multi-view DL network for ATR by constructing a parallel network topology with multiple inputs. On the basis of adversarial learning, Guo \textit{et al}. \cite{r49} developed a unified SAR ATR framework integrating data denoising, feature extraction and recognition. Given the evident discrepancies between optical and SAR images, many researchers have attempted to combine the unique attribute scattering center features and convolutional neural networks (CNNs) for ATR. A feature fusion scheme was presented by Zhang \textit{et al}. \cite{r23}, for integrating semantic features extracted by a CNN and target scattering centers. Discriminant correlation analysis, which maximized the correlation between CNNs and attribute scattering centers, was presented for feature fusion in the framework. Qin \textit{et al}. \cite{r13} developed a scattering attribute embedded network for the small sample SAR ATR, which exploited local structure and global semantic representation to incorporate the inherent structural attributes of SAR targets. Hou \textit{et al}. \cite{r24} introduced a scattering centers-integrated graph neural network for SAR ATR. The scattering center features were added to the training process in the form of graphs, and the structural features and physical information were extracted and learned through graph neural networks with feature-wise linear modulation.

\subsection{Simulated data-assisted SAR ATR}
SAR image simulation technology has the advantages of high flexibility and low cost, providing large-scale annotated synthetic datasets for training the DL-based models. In recent years, simulated data-assisted SAR ATR has emerged as a promising research area.

Malmgren-Hansen \textit{et al}. \cite{r53} addressed the viability of transfer learning between the real and synthetic SAR images initially. The experimental results demonstrated that, in situations where the labeled real samples were insufficient, pretraining with the simulated data could successfully boost its classification accuracy. An adversarial coding network was proposed by Du \textit{et al}. \cite{r26} to extract shared features between measured and simulated data. In order to align the global simulation and measurement distributions, Shi \textit{et al}. \cite{r15} presented an adversarial learning strategy by weighting gradients. They also looked into a category-level alignment technique by using a prototype network with fine-grained class structures. Lv \textit{et al}. \cite{r28} employed a combination of subdomain alignment and dual branching image reconstruction to effectively narrow the domain gap. In light of subclasses domain shifting and special SAR image feature mining, Zhang \textit{et al}. \cite{r29} presented a domain adaption network with a two-stage adaptive loss, merging scattering topological and visual features. Youk \textit{et al}. \cite{r54} designed a transformer-based network to translate a simulated SAR image to a measured version. For synthetic data-aided SAR ATR, Han \textit{et al}. \cite{r47} established a trusted knowledge distillation network, by assessing the effectiveness of knowledge learned from synthetic data.

\subsection{Semi-supervised Domain Adaptation}
The fact that the target domain samples lack labels severely hinders cross-domain recognition. Several investigations have started to concentrate on SSDA, which shows strong cross-domain recognition performance for optical image classification tasks.

Kuniaki \textit{et al}. \cite{r30} proposed a minimum maximum entropy (MME) method that adversarially optimized an adaptive less labeled model. Taekyung \textit{et al}. \cite{r31} presented a framework for SSDA that aligned features by reducing domain differences, consisting of three main techniques, namely attraction, perturbation and exploration. Li \textit{et al}. \cite{r33} created a domain alignment technique based on enhanced categorical alignment united with consistency learning. By combining self-training and maximum mean difference distance minimization, Qin \textit{et al}. \cite{r34} introduced an adaptive structure learning approach that addresses cross-domain alignment as well as limited labeled data learning. Aiming to accomplish class alignment, Li \textit{et al}. \cite{r35} introduced an SSDA technique termed graph-based adaptive betweeness clustering, which imposed semantic transfer to unlabeled target instances from labeled instances. An SSDA model was presented by Yu \textit{et al}. \cite{r36} to adapt the source samples by changing their labels to the target space. The main concept lay in that the source instances could be regarded as a noisy-labeled version in the target space.

However, there are few SSDA studies for SAR ATR using simulated data, despite the fact that SSDA investigations are relatively common in optical image recognition. 

\begin{figure*}[htbp]
	\centering
	\includegraphics[width=1.0\textwidth]{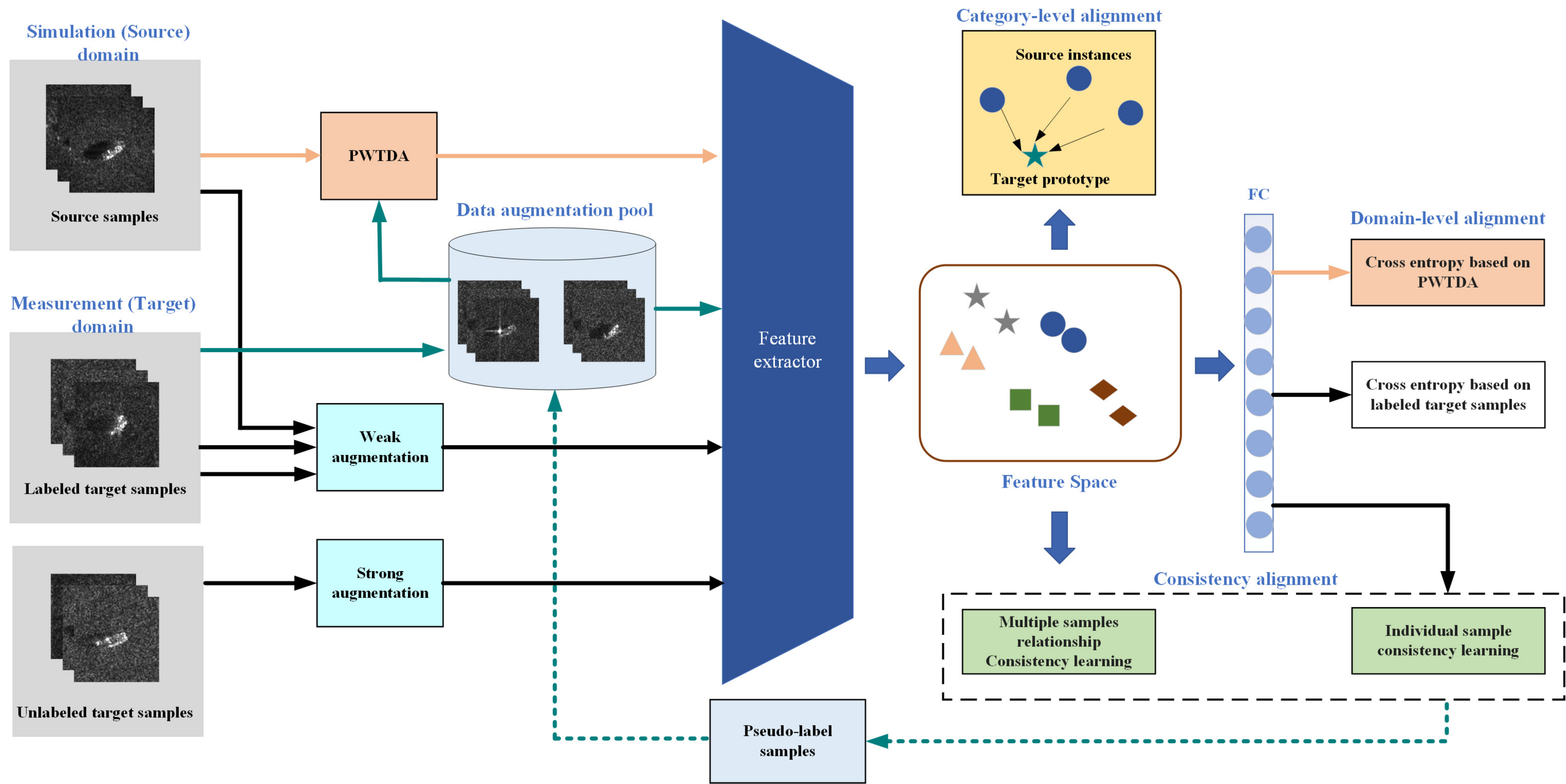}
	\caption{A synopsis of the suggested SSDA method. There are three components to this method: domain alignment, category alignment and consistency alignment. Based on the PWTDA, the domain alignment is accomplished. The category alignment is obtained by instance-prototype alignment between the simulated and measured domains. Furthermore, consistency alignment is exclusively applied to the measured domain, implemented by individual sample consistency learning and multi-sample relationship consistency alignment.}
	\label{fig_1}
\end{figure*}

\section{Method}
\label{sec3}
SSDA studies the transfer of knowledge from a labeled source domain ${\cal S} = \left\{ {\left( {{\boldsymbol{s}_i},y_i^s} \right)} \right\}_{i = 1}^{{N_s}}$ to a target domain  ${\cal T} = {\cal L} \cup {\cal U}$ with only partial labels, where $y_i^s$ is the label of source sample ${\boldsymbol{s}_i}$ and ${N_s}$ is the quantity of source samples. ${\cal L} = \left\{ {\left( {{\boldsymbol{l}_i},y_i^l} \right)} \right\}_{i = 1}^{{N_l}}$ is a labeled set, where $y_i^l$ is the label of labeled target image ${\boldsymbol{l}_i}$ and ${N_l}$ is the number of labeled target images. And ${\cal U} = \left\{ {\left( {{\boldsymbol{u}_i}} \right)} \right\}_{i = 1}^{{N_u}}$ is an unlabeled set, where ${\boldsymbol{u}_i}$ is a labeled target sample and ${N_u}$ represents the quantity of unlabeled target samples. The labels of ${\cal S}$ and ${\cal T}$ both originate from a label space ${\cal Y} = \left\{ {1,2, \ldots C} \right\}$. In this paper, the Synthetic and Measured Paired Labeled Experiment (SAMPLE) dataset is the main experimental dataset, where the simulated dataset is treated as ${\cal S}$ and the measured dataset as ${\cal T}$ \cite{r14,r37}. Apparently, data distributions of ${\cal S}$ and ${\cal T}$ are different, leading to domain shifting. Both ${\cal S}$ and ${\cal T}$ are leveraged to learn a domain adaptation model denoted as $h = f \circ g$ with parameters $\theta $, in which $f$ stands for the feature extractor and $g$ for the classifier. Typically, regarding SSDA, labeled samples in ${\cal T}$ is quite scarce, bringing huge challenges for the domain adaptation. According to Fig. \ref{fig_1}, the proposed SSDA framework considers domain alignment, category alignment and consistency alignment, in which the key modules consist of the PWTDA, AIPA and consistency learning. The PWTDA encourages the simulated domain distribution to approach the measured domain distribution, aiming to achieve domain-level alignment. The AIPA facilitates the simulated domain samples to approximate to the measured prototypes of the corresponding class, so as to attain category alignment. The consistency learning includes single-sample strong-weak augmentation consistency and multi-sample relationship consistency. In this paper, there are four parts in the comprehensive loss: the supervised loss, the progressive prototype alignment loss, the pseudo-label loss and the multi-sample relationship consistency loss. The supervised loss is associated with the PWTDA, and the progressive prototype alignment loss with the AIPA. The pseudo-label loss and the multi-sample relationship consistency loss force the model to implement consistency alignment by taking advantage of measured domain samples without labels. Each module is described as follows.

\begin{figure}[!t]
	\centering
	\includegraphics[width=0.50\textwidth]{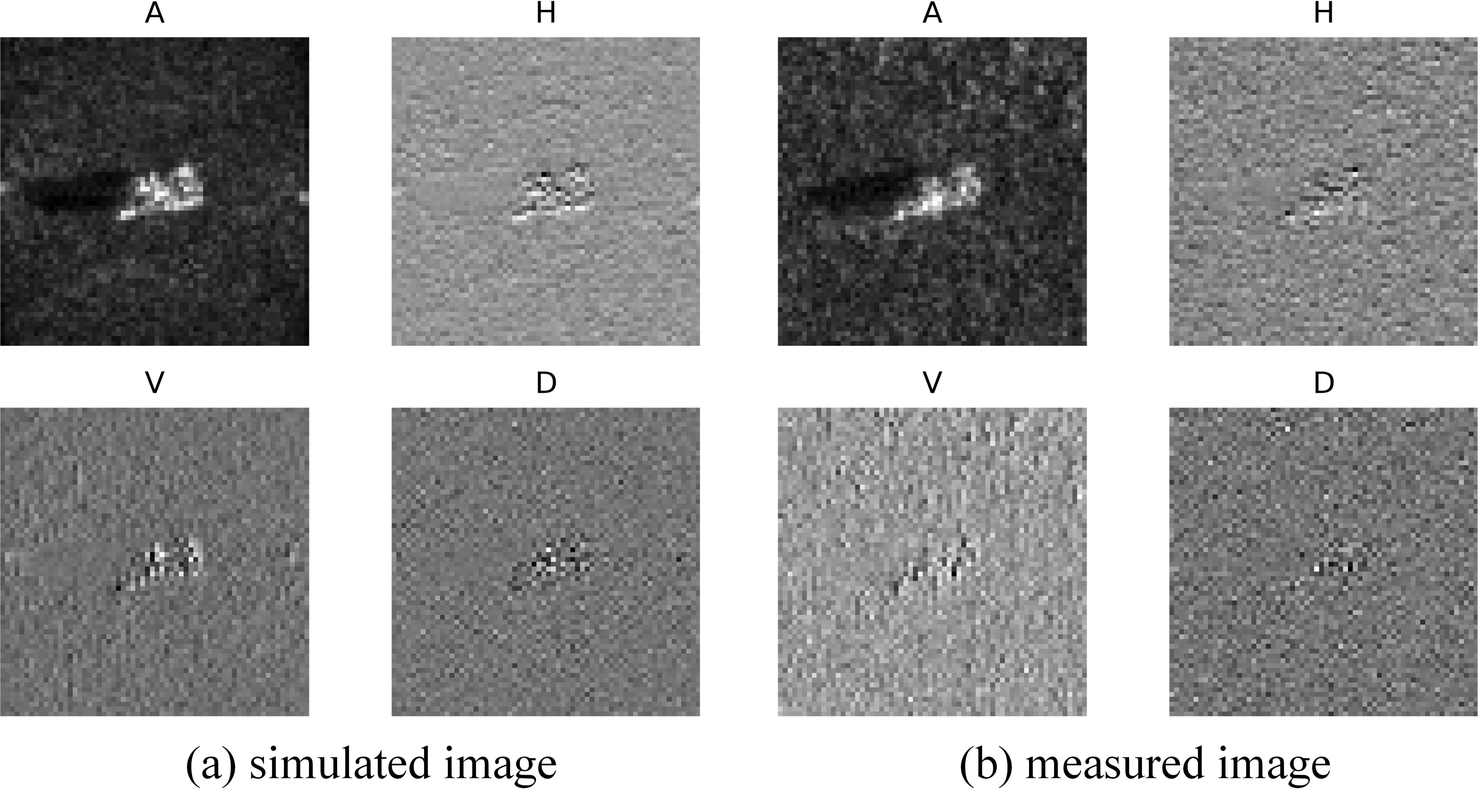}
	\caption{An example of wavelet decompoisition of a simulated-measured SAR image pair.}
	\label{fig_2}
\end{figure}

\subsection{Progressive Wavelet Transform Data Augmentation}
\label{sec3-A}
Data augmentation is usually adopted as a critical technique for domain adaptation. Some studies demonstrated that frequency domain data augmentation can effectively enhance the domain generalization capability \cite{r20,r21}. Xu \textit{et al}. \cite{r20} proposed a method for data augmentation in the frequency domain by obfuscating the amplitude spectrum information of samples. In contrast with the Fourier transform, the wavelet transform permits a more precise local description as well as the extraction of signal features \cite{r38,r39,r40}. This paper presents PWTDA, a method that utilizes high-frequency sub-band mixing of wavelet-transformed images. 

\subsubsection{Wavelet Transform Data Augmentation}
For a single SAR image ${\rm I} \left( {i,j} \right)$, its 2D discrete wavelet decomposition is obtained using the Mallat algorithm.
\begin{figure*}[!t]
	\centering
	\includegraphics[width=0.90\textwidth]{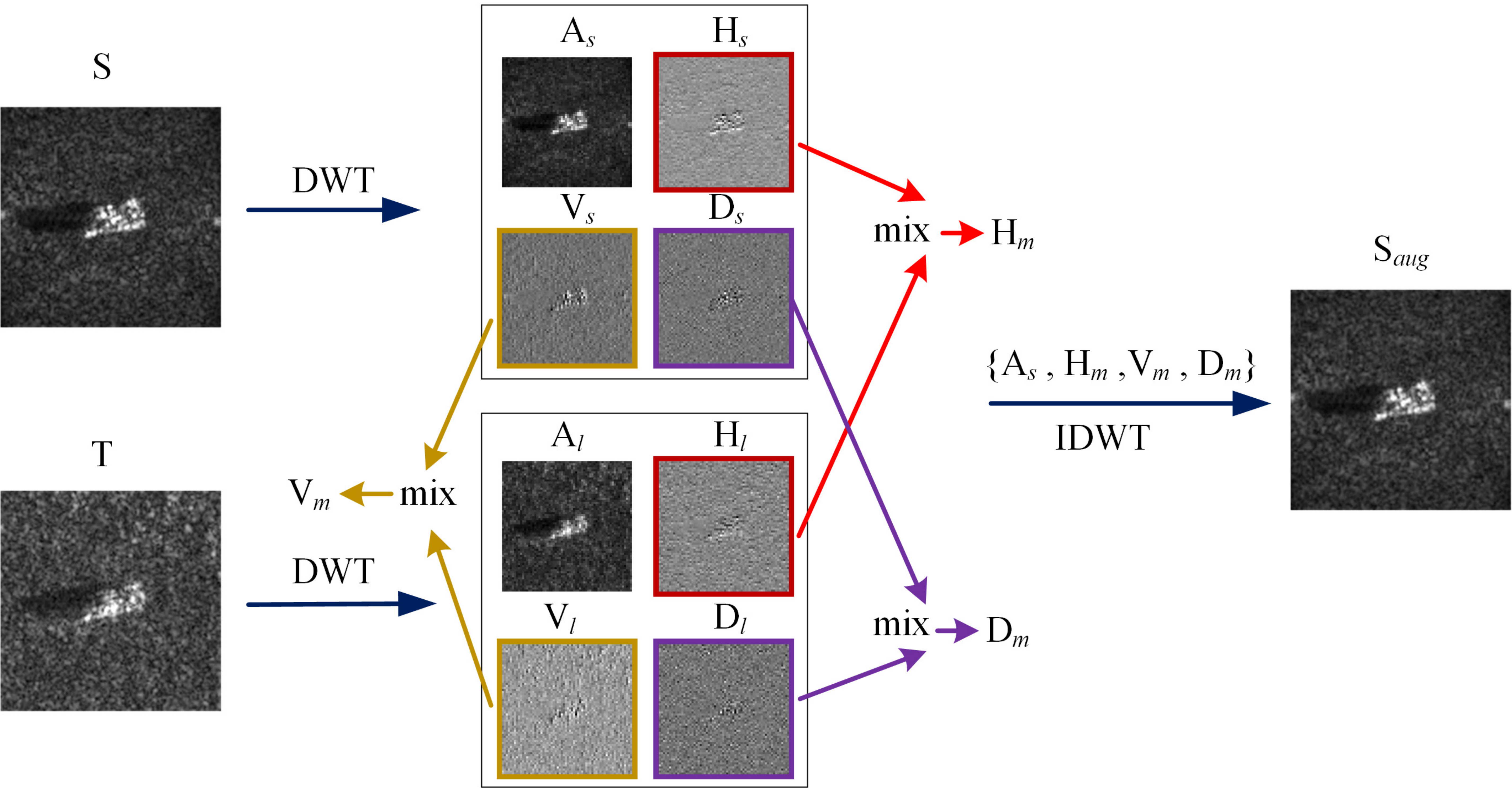}
	\caption{Schematic diagram of the wavelet transform data augmentation.}
	\label{fig_3}
\end{figure*}

\begin{equation}
	\label{eq1}
	{\rm A}(m,n) = \sum\limits_{i,j} {{\rm I}\left( {i,j} \right){{\tilde L}_F}\left( {i - 2m} \right){{\tilde L}_F}\left( {j - 2n} \right)} 
\end{equation}
\begin{equation}
	\label{eq2}
	{\rm H}(m,n) = \sum\limits_{i,j} {{\rm I}\left( {i,j} \right){{\tilde L}_F}\left( {i - 2m} \right){{\tilde H}_F}\left( {j - 2n} \right)} 
\end{equation}
\begin{equation}
	\label{eq3}
	{\rm V}(m,n) = \sum\limits_{i,j} {{\rm I}\left( {i,j} \right){{\tilde H}_F}\left( {i - 2m} \right){{\tilde L}_F}\left( {j - 2n} \right)} 
\end{equation}
\begin{equation}
	\label{eq4}
	{\rm D}(m,n) = \sum\limits_{i,j} {{\rm I}\left( {i,j} \right){{\tilde H}_F}\left( {i - 2m} \right){{\tilde H}_F}\left( {j - 2n} \right)} 
\end{equation}
where ${\tilde H_F}$ stands for the high-pass and ${\tilde L_F}$ for low-pass filters. The wavelet transform decomposes the SAR image into a low-frequency sub-band (denoted by A) and three high-frequency sub-bands (denoted by H, V, and D) in \eqref{eq1}, \eqref{eq2}, \eqref{eq3}, \eqref{eq4}. The A component usually represents the contour of the image. Additionally, the horizontal, vertical, and diagonal directions of the image's detailed information are included in H, V, and D \cite{r40}. For the image $\rm I$, the symbol “${\rm{DWT2}}(\cdot )$” is used to represent the 2D discrete wavelet decomposition in \eqref{eq5}:
\begin{equation}
	\label{eq5}
	\rm A,H,V,D = {\rm{DWT2}}(I)
\end{equation}

The four sub-bands A, H, V, and D, can be used to rebuild the original image, by inverse wavelet transform in \eqref{eq6}. ${H_F}$ stands for the high-pass and ${L_F}$ for low-pass filters. Here, the symbol “${\rm{IDWT2}}(\cdot )$” denotes the 2D discrete wavelet reconstruction in \eqref{eq7}.
\begin{equation}
\label{eq6}
	\begin{array}{l}
		{\rm I}\left( {i,j} \right) = \sum\limits_{m,n} {{\rm A}\left( {m,n} \right){L_F}\left( {m - 2i} \right){L_F}\left( {n - 2j} \right)} \\
		{\kern 1pt} {\kern 1pt} {\kern 1pt} {\kern 1pt} {\kern 1pt} {\kern 1pt} {\kern 1pt} {\kern 1pt} {\kern 1pt} {\kern 1pt} {\kern 1pt} {\kern 1pt} {\kern 1pt} {\kern 1pt} {\kern 1pt} {\kern 1pt} {\kern 1pt} {\kern 1pt} {\kern 1pt} {\kern 1pt} {\kern 1pt} {\kern 1pt} {\kern 1pt} {\kern 1pt} {\kern 1pt} {\kern 1pt} {\kern 1pt} {\kern 1pt} {\kern 1pt} {\kern 1pt} {\kern 1pt}  + \sum\limits_{m,n} {{\rm H}\left( {m,n} \right){L_F}\left( {m - 2i} \right){H_F}\left( {n - 2j} \right)} \\
		{\kern 1pt} {\kern 1pt} {\kern 1pt} {\kern 1pt} {\kern 1pt} {\kern 1pt} {\kern 1pt} {\kern 1pt} {\kern 1pt} {\kern 1pt} {\kern 1pt} {\kern 1pt} {\kern 1pt} {\kern 1pt} {\kern 1pt} {\kern 1pt} {\kern 1pt} {\kern 1pt} {\kern 1pt} {\kern 1pt} {\kern 1pt} {\kern 1pt} {\kern 1pt} {\kern 1pt} {\kern 1pt} {\kern 1pt} {\kern 1pt} {\kern 1pt} {\kern 1pt} {\kern 1pt} {\kern 1pt}  + \sum\limits_{m,n} {{\rm V}\left( {m,n} \right){H_F}\left( {m - 2i} \right){L_F}\left( {n - 2j} \right)} \\
		{\kern 1pt} {\kern 1pt} {\kern 1pt} {\kern 1pt} {\kern 1pt} {\kern 1pt} {\kern 1pt} {\kern 1pt} {\kern 1pt} {\kern 1pt} {\kern 1pt} {\kern 1pt} {\kern 1pt} {\kern 1pt} {\kern 1pt} {\kern 1pt} {\kern 1pt} {\kern 1pt} {\kern 1pt} {\kern 1pt} {\kern 1pt} {\kern 1pt} {\kern 1pt} {\kern 1pt} {\kern 1pt} {\kern 1pt} {\kern 1pt} {\kern 1pt} {\kern 1pt} {\kern 1pt}  + \sum\limits_{m,n} {{\rm D}\left( {m,n} \right){H_F}\left( {m - 2i} \right){H_F}\left( {n - 2j} \right)} 
	\end{array}
\end{equation}
\begin{equation}
	\label{eq7}
	\rm I = IDWT2(A,H,V,D)
\end{equation}

We analyzed the wavelet transform results for a substantial portion of the SAMPLE dataset’s samples. It is found that the primary discrepancy between the simulated and measured image pairs exists in the high-frequency parts rather than low-frequency parts. The phenomena could be observed in Fig. \ref{fig_2}, where the decomposition results of an image pair are shown schematically. This is because the simulated images are calculated based on the target CAD model and computational EM methods \cite{r14,r37}. Due to inaccuracies in the CAD modeling and EM approximation, there are certain limitations in the high-frequency detail information.

Based on the aforementioned analysis, the model should narrow domain gap from the high-frequency sub-band information so as to obtain an excellent cross-domain generalization ability. To this end, a natural choice is to perturb the high-frequency parts information of the source samples by those of target samples. Motivated by Mixup \cite{r41}, we designed a wavelet transform data augmentation method confusing the high-frequency sub-bands of images from both target and source domains. The high-frequency parts are mixed, and then united with the original low-frequency parts of source domain images to generate the augmented images via IDWT2. The implementation procedure is as follows. Apply DWT2 to decompose a simulated domain image ${\boldsymbol{s}_i}$ and a labeled measured domain image ${\boldsymbol{l}_i}$ of the corresponding class in \eqref{eq8}, \eqref{eq9}.
\begin{equation}
	\label{eq8}
	{{\rm A}_s},{{\rm H}_s},{{\rm V}_s},{{\rm D}_s} = {\rm DWT2}({\boldsymbol{s}_i})
\end{equation}
\begin{equation}
	\label{eq9}
	{{\rm A}_l},{{\rm H}_l},{{\rm V}_l},{{\rm D}_l} = {\rm DWT2}({\boldsymbol{l}_i})
\end{equation}

And mix their high-frequency sub-bands with a certain weight in \eqref{eq10}, \eqref{eq11}, \eqref{eq12}. 
\begin{equation}
	\label{eq10}
	{{\rm H}_m} = \alpha {{\rm H}_s} + (1 - \alpha ){{\rm H}_l}
\end{equation}
\begin{equation}
	\label{eq11}
	{{\rm V}_m} = \alpha {{\rm V}_s} + (1 - \alpha ){{\rm V}_l}
\end{equation}
\begin{equation}
	\label{eq12}
	{{\rm D}_m} = \alpha {{\rm D}_s} + (1 - \alpha ){{\rm D}_l}
\end{equation}
where $\alpha $ is the hyperparameter of the mixing ratio. Finally, combine them with the low-frequency part of ${\boldsymbol{s}_i}$ to form a new sample by IDWT2, producing the augmented version of ${\boldsymbol{s}'_i}$ in \eqref{eq13}. The principle of the wavelet transform data augmentation is shown in Fig. \ref{fig_3} and represented in \eqref{eq8}, \eqref{eq9}, \eqref{eq10}, \eqref{eq11}, \eqref{eq12}, \eqref{eq13}. 
\begin{equation}
	\label{eq13}
	{\mathbf{s}'_i} = {\rm IDWT2}({{\rm A}_s},{{\rm H}_m},{{\rm V}_m},{{\rm D}_m})
\end{equation}

\begin{figure*}[htbp]
	\centering
	\includegraphics[width=0.90\textwidth]{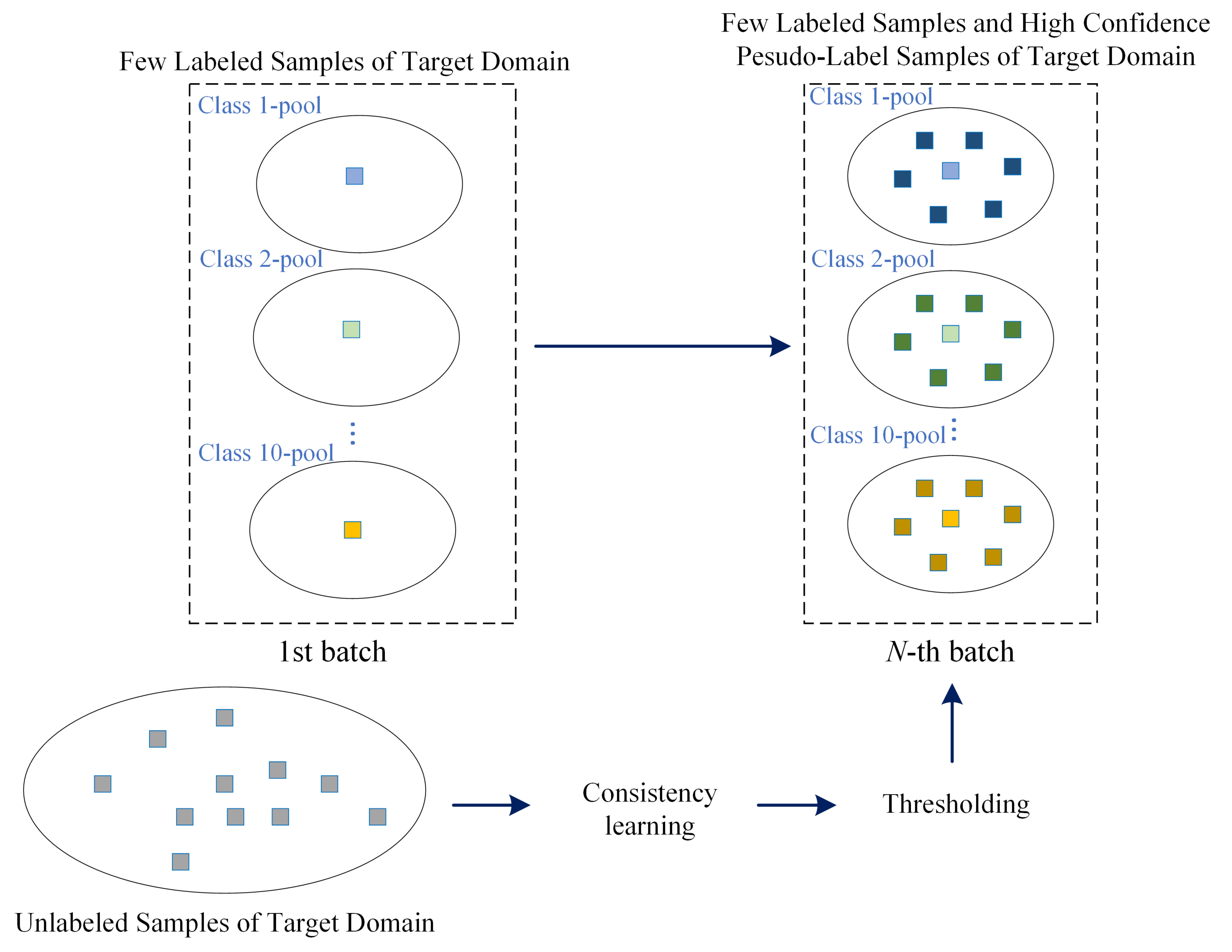}
	\caption{Schemetic diagram of the data augmentation pools.}
	\label{fig_4}
\end{figure*}

It is worth noting that the high-frequency sub-bands are blended rather than low-frequency sub-bands. The primary causes are as follows in two parts. On the one hand, the azimuth angle has an impact on the measured image \cite{r15}. If the azimuth angle of the target sample  differs greatly from that of source sample, the great discrepancy will emerge between their low-frequency parts. Thus, mixing their low-frequency parts will lead to severe information confusion and unmeaningful augmented samples. On the other hand, the measured domain contains just a few of labeled samples, lacking diversity. Under these circumstances, if low-frequency sub-bands of two domain samples are mixed, low-frequency information will play a smaller part in the blended source domain data, resulting in deterioration of the source samples diversity and the model discrimination ability. In addition, our method perturbs each high-frequency sub-band, and brings a more generalized image by weighted linear blending. As a result, by contrasting the images before and after augmentation, the model is able to learn from the high-frequency sub-bands information in an efficient manner.

\subsubsection{Progressive augmentation strategy}
Owing to the target domain’s dearth of labeled samples, few target samples are involved in the augmentation of source samples. Hence, if the target domain is limited to a small amount of labeled instances, the augmented sample diversity is lower. Moreover, it is well known that SAR target image characteristics are closely related to imaging azimuth angles. SAR images with different azimuth angles may present enormous discrepancy in information. The model should acquire target knowledge from multiple azimuth samples as much as possible. Therefore, to make the augmented samples possess predominant diversity and generalization, more target domain samples with various azimuth angles need to be involved in the data augmentation.

\begin{figure*}[!t]
	\centering
	\includegraphics[width=1.0\textwidth]{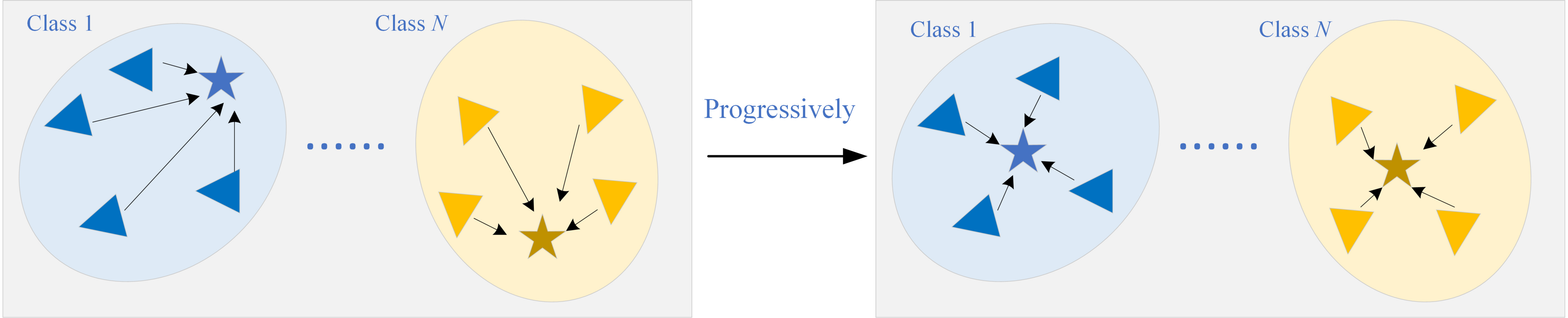}
	\caption{Schematic diagram of the AIPA. As the data augmentation pools expand, every target prototype approaches the relevant real class distribution more closely. The characteristics of the source domain samples are represented by the triangles, and the pentagrams indicate the class target prototypes. The same color represents the same class.}
	\label{fig_5}
\end{figure*}

To this end, we present the progressive wavelet transform augmentation strategy by utilizing unlabeled target samples to the fullest. The task involves predicting pseudo-labels for the unlabeled samples in the target domain. As the model is being trained, the data augmentation is continuously added high-confidence pseudo-labeled examples. A data augmentation pool ${\cal M} = \left\{ {\left( {{\boldsymbol{m} _i},y_i^m} \right)} \right\}_{i = 1}^{{N_m}}$, where $y_i^m$ is the label of the image ${\boldsymbol{m}_i}$ participating in the wavelet data augmentation and ${N_m}$ is the number of target images of each class. In ${\cal M}$, the labeled samples ${\boldsymbol{l}_i}$ of the class are stored as well as high-confidence pseudo-label samples $\boldsymbol{u}_i^w\left( {\max \left( {h\left( {\boldsymbol{u}_i^w} \right)} \right) \ge \sigma } \right)$ (unlabeled samples whose highest confidence exceeds the likelihood threshold $\sigma $) in \eqref{eq14}. For how to obtain the high-confidence pseudo-label samples, please refer to the Section \ref{sec3-C}. To improve the number of instances, the data augmentation pools are additionally supplemented with weakly augmented versions of high-confidence unlabeled samples. At the beginning of the model training, each data augmentation pool contains very few labeled target samples. Along with the model training, the data augmentation pool is added with reliable unlabeled instances from the current batch. At the next batch, all samples in each data augmentation pool are used for the wavelet transform data augmentation. In this way, with the increase of training batch, there are more and more high-confidence pseudo-labeled samples in these pools. Thus, more pseudo-labeled target samples take part in the augmentation process, enhancing both the model’s generalization and sample diversity. Fig. \ref{fig_4} presents an illustration of the data augmentation pools.
\begin{equation}
	\label{eq14}
	{\cal M} = \left\{ {{\boldsymbol{l}_i}\& \boldsymbol{u}_i^w\left( {\max \left( {h\left( {\boldsymbol{u}_i^w} \right)} \right) \ge \sigma } \right)} \right\}
\end{equation}

As the supervised loss, the cross entropy is used for loss calculation for both augmentation source samples and labeled target samples.

\subsection{Asymptotic Instance-Prototype Alignment}
In SSDA, finer class-level alignment is more likely to be achieved because target samples with labels are available. To implement class-level alignment, we developed an instance-prototype alignment approach. Assuming that a prototype in the feature space represents a corresponding target class, the distance between each target prototype and the source domain samples is calculated to get the recognition results \cite{r15}. Note that it is instance-prototype alignment, rather than prototype-prototype alignment. When training the model, the instance-prototype loss is computed using source samples and target prototypes, with the goal of bringing the former closer to their respective target category prototypes. In \eqref{eq15}, by calculating the feature embedding $f\left( {{\boldsymbol{l}_i}} \right)$ of the labeled target samples ${\boldsymbol{l}_i}$, we obtain the target prototypes $\left\{ {{\boldsymbol{h}_k}} \right\}_{k = 1}^C$ for every class \cite{r42}.
\begin{equation}
	\label{eq15}
	{\boldsymbol{h}_k} = \frac{1}{{\left| {{{\rm{{\cal T}}}_k}} \right|}}\sum\limits_{\left( {{\boldsymbol{l}_i}y_i^l} \right) \in {{\rm{{\cal T}}}_k}} {f\left( {{\boldsymbol{l}_i}} \right)} 
\end{equation} 

Then, in \eqref{eq16}, the classification probability of the source sample is calculated based on each source domain sample’s distance from every target prototype \cite{r33}.
\begin{equation}
	\label{eq16}
	p\left( {y_i^s = {y^s}|{\boldsymbol{s}_i}} \right) = \frac{{\exp \left( { - {{\left\| {f\left( {{\boldsymbol{s}_i}} \right) - {\boldsymbol{h}_y}} \right\|}_2}} \right)}}{{\sum\nolimits_{k = 1}^C {\exp \left( { - {{\left\| {f\left( {{\boldsymbol{s}_i}} \right) - {\boldsymbol{h}_k}} \right\|}_2}} \right)} }}
\end{equation}

Finally, we compute the instance-prototype loss ${L_{pta}}$ on all source samples in \eqref{eq17}. Evidently, minimizing the instance-prototype loss achieves cross-domain class-level alignment.
\begin{equation}
	\label{eq17}
	{L_{pta}} = \frac{1}{{C{N_s}}}\sum\limits_{\left( {{\boldsymbol{s}_i},y_i^s} \right) \in {\cal S}} {\sum\limits_{y \in {\cal Y}} {y_i^s\log \left[ { - p\left( {y_i^s = y^s|{\boldsymbol{s}_i}} \right)} \right]} } 
\end{equation}

However, because there aren’t sufficient target samples with labels, the class prototypes obtained based on labeled samples cannot accurately reflect the real distribution of each class. To overcome this issue, similar to the PWTDA, the high-confidence unlabeled samples are also incorporated into the computation of the target prototypes. That is, the samples ${\boldsymbol{m}_i}$ in the data augmentation pool ${\cal M} = \left\{ {\left( {{\boldsymbol{m}_i},y_i^m} \right)} \right\}_{i = 1}^{{N_m}}$ are used to compute the target prototypes in \eqref{eq15}. With the increase of the samples in the data augmentation pools, the updated target prototypes gradually approximate the true distributions. Therefore, AIPA has a significant role in class-level alignment, the principle of which is shown in Fig. \ref{fig_5}. 

\subsection{Consistency alignment}
\label{sec3-C}
Further improving the model’s capacity for generalization, we incorporate weak-strong augmentation consistency learning into the proposed approach for SAR ATR. It has been shown that consistency regularization is a useful technique for learning from unlabeled data \cite{r43}. In essence, for unlabeled data, weak augmentation is used to create pseudo-labels. These pseudo-labels are then combined with strong augmentation to determine the cross-entropy loss \cite{r22}. The model’s robustness to perturbation of the input samples is boosted by this constant regularization, which concretely conducts the consistency alignment between the labeled and unlabeled target samples. In this paper, the consistency learning process involves all unlabeled target samples. Particularly, two type of consistency are considered, including single sample and multi-sample relationship consistency. 

To excavate the single sample consistency, for a target domain unlabeled sample ${\boldsymbol{u}_i} \in {\cal U}$, we obtain the weak augmentation version $\boldsymbol{u}_i^w$ and strong augmentation version $\boldsymbol{u}_i^s$ followed by the augmentation technique in \cite{r44}. Feed $\boldsymbol{u}_i^w$ into the model, and obtain its classification probabilities ${\boldsymbol{p}_w}$. The same operation is applied to $\boldsymbol{u}_i^s$, gaining ${\boldsymbol{p}_s}$. Afterward, the pseudo-labels ${\tilde {\boldsymbol{p}}_w}$ are obtained based on ${\boldsymbol{p}_w}$ according to \eqref{eq18}. 
\begin{equation}
	\label{eq18}
	{\tilde {\boldsymbol{p}}_w} = \arg \max \left( {{\boldsymbol{p}_w}} \right)
\end{equation}

The pseudo-label loss ${L_{pl}}$ \cite{r17} can be calculated by
\begin{equation}
	\label{eq19}
	{L_{pl}} = \sum\limits_{{\boldsymbol{u}_i} \sim {\cal U}} {\left[ {1\left( {\max \left( {{p_w}} \right) \ge \sigma } \right)H\left( {{{\tilde {\boldsymbol{p}}}_w},{\boldsymbol{p}_s}} \right)} \right]} 
\end{equation}
where the cross-entropy of ${\tilde {\boldsymbol{p}}_w}$ and ${\boldsymbol{p}_s}$ is represented by $H\left( {:,:} \right)$; $1\left( {\max \left( {{\boldsymbol{p}_w}} \right) \ge \sigma } \right)$ indicates that only samples with the highest probability exceeding the threshold $\sigma $ are counted. 

For exploiting multiple samples relationship consistency, the group relationship consistency loss in \cite{r10} is introduced for the model training. Use ${{\mathbf{F}}_{s}}$ and ${{\mathbf{F}}_{w}}$ to indicate the feature set of the strongly and weakly augmented samples respectively. Firstly, any two feature vectors within ${{\mathbf{F}}_{s}}$ have their similarity computed by Gaussian radial basis function as in \eqref{eq20}. Then construct the similarity matrix ${{\mathbf{H}}_{s}}$ as \eqref{eq21}. The same operation is applied to ${{\mathbf{F}}_{w}}$, gaining ${{\mathbf{H}}_{w}}$.
\begin{equation}
	\label{eq20}
	{H_{ij}} = \exp \left( { - \frac{{{{\left\| {f\left( {{\boldsymbol{u}_i}} \right) - f\left( {{\boldsymbol{u}_j}} \right)} \right\|}^2}}}{{2{\beta ^2}}}} \right) 
\end{equation}
where ${H_{ij}}$ denotes the similarity between $f\left( {{\boldsymbol{u}_i}} \right)$ and $f\left( {{\boldsymbol{u}_j}} \right)$, ${H_{ij}} \in \left[ {0,1} \right]$. In the embedding space, the value ${H_{ij}}$ represents the level of similarity between two samples. $\beta $ is the hyperparameter. 
\begin{equation}
	\label{eq21}
	\mathbf{H} = \left[ {\begin{array}{*{20}{c}}
			{{H_{11}}}&{{H_{12}}}& \cdots &{{H_{1{N_U}}}}\\
			{{H_{21}}}&{{H_{22}}}& \cdots &{{H_{2{N_U}}}}\\
			\vdots & \vdots & \cdots & \vdots \\
			{{H_{{N_U}1}}}&{{H_{{N_U}2}}}& \cdots &{{H_{{N_U}{N_U}}}}
	\end{array}} \right]
\end{equation}

${\mathbf{H}_s}$ indicates the relationship between strong augmented versions of all unlabeled target data, and ${\mathbf{H}_w}$ presents the relationship between weak augmentations. ${L_{msr}}$ is computed by the mean square error of ${\mathbf{H}_s}$ and ${\mathbf{H}_w}$ in \eqref{eq21}.
\begin{equation}
	\label{eq22}
	{L_{msr}} = \frac{1}{{{N_U}^2}}\sum\limits_{i = 1}^{{N_U}} {\sum\limits_{j = 1}^{{N_U}} {\left\| {{\mathbf{H}_s}\left( {i,j} \right) - {\mathbf{H}_w}\left( {i,j} \right)} \right\|_2^2} } 
\end{equation}
where the distance between ${\mathbf{H}_s}$ and ${\mathbf{H}_w}$ is indicated by ${L_{msr}}$. By optimally designing the loss term, the model is encouraged to learn global semantic knowledge.

Finally, the overall consistency alignment loss is as follows.
\begin{equation}
	\label{eq23}
	{L_{cona}} = {L_{pl}} + {\lambda _{msr}}{L_{msr}}
\end{equation}
where ${\lambda _{msr}}$ denotes the weight parameter, that balances the pseudo-label loss with the multi-sample relationship loss.

\begin{algorithm}[t]
\renewcommand{\algorithmicrequire}{\textbf{Input:}}
\caption{Training Process.}\label{alg:alg1}
\begin{algorithmic}
\STATE 
\REQUIRE  The source domain ${\cal S} = \left\{ {\left( {{\boldsymbol{s}_i},y_i^s} \right)} \right\}_{i = 1}^{{N_s}}$; the labeled target domain ${\cal L} = \left\{ {\left( {{\boldsymbol{l}_i},y_i^l} \right)} \right\}_{i = 1}^{{N_l}}$; the unlabeled target domain ${\cal U} = \left\{ {\left( {{\boldsymbol{u}_i}} \right)} \right\}_{i = 1}^{{N_u}}$.

\STATE $\mathbf{step 1}$: Strong and weak augmentation of raw data. Weakly augment the labeled target samples and source samples. Both weakly and strongly, the unlabeled target samples are augmented. 
\STATE $\mathbf{step 2}$: Data augmentation pool ${\cal M}$ construction and initialization: ${\cal M} = {\cal L}$; ${\cal M} = \left\{ {\left( {{\boldsymbol{m}_i},y_i^m} \right)} \right\}_{i = 1}^{{N_m}}$;
\STATE $\mathbf{step 3}$: Acquire the wavelet-domain augmented version of the source-domain data in  \eqref{eq8}, \eqref{eq9}, \eqref{eq10}, \eqref{eq11}, \eqref{eq12}, \eqref{eq13};
\STATE $\mathbf{step 4}$: Compute cross-entropy loss ${L_{wte}}$ for the wavelet-domain augmented version of
the source-domain data;
\STATE $\mathbf{step 5}$: Calculate target prototypes based on the data augmentation pool ${\cal M}$ through \eqref{eq15};
\STATE $\mathbf{step 6}$: Calculate prototype losses ${L_{pta}}$ via \eqref{eq16}, \eqref{eq17};
\STATE $\mathbf{step 7}$: Calculate pseudo-label loss ${L_{pl}}$ and multi-sample relationship loss ${L_{msr}}$ by \eqref{eq19} and \eqref{eq22};
\STATE $\mathbf{step 8}$: Add pseudo-label data $\boldsymbol{u}_i^w\left( {\max \left( {h\left( {\boldsymbol{u}_i^w} \right)} \right) \ge \sigma } \right)$ to the data augmentation pool ${\cal M}$;
\STATE $\mathbf{step 9}$: Obtain the total loss with \eqref{eq24}. Jump to Step 3 if unlabeled data is not calculated; jump to Step 2 if it is; repeat;
\end{algorithmic}
\label{alg1}
\end{algorithm}

\subsection{Total loss}
In summary, the following is the entire loss function.
\begin{equation}
	\label{eq24}
	L = {L_{wte}} + {\lambda _{pta}}{L_{pta}} + {\lambda _{cona}}{L_{cona}}
\end{equation}
where ${L_{wte}}$ is the cross-entropy of the wavelet transform augmentation of the source domain sample, ${\lambda _{pta}}$ and ${\lambda _{cona}}$ denote the weight parameters of the instance-prototype alignment and the consistency alignment loss term respectively. Algorithm \ref{alg1} explains how our suggested technique is trained.

\section{Numerical Experiments}
\label{sec4}
\subsection{Introduction to the dataset and experimental setup}
\begin{figure}[!t]
	\centering
	\includegraphics[width=0.5\textwidth]{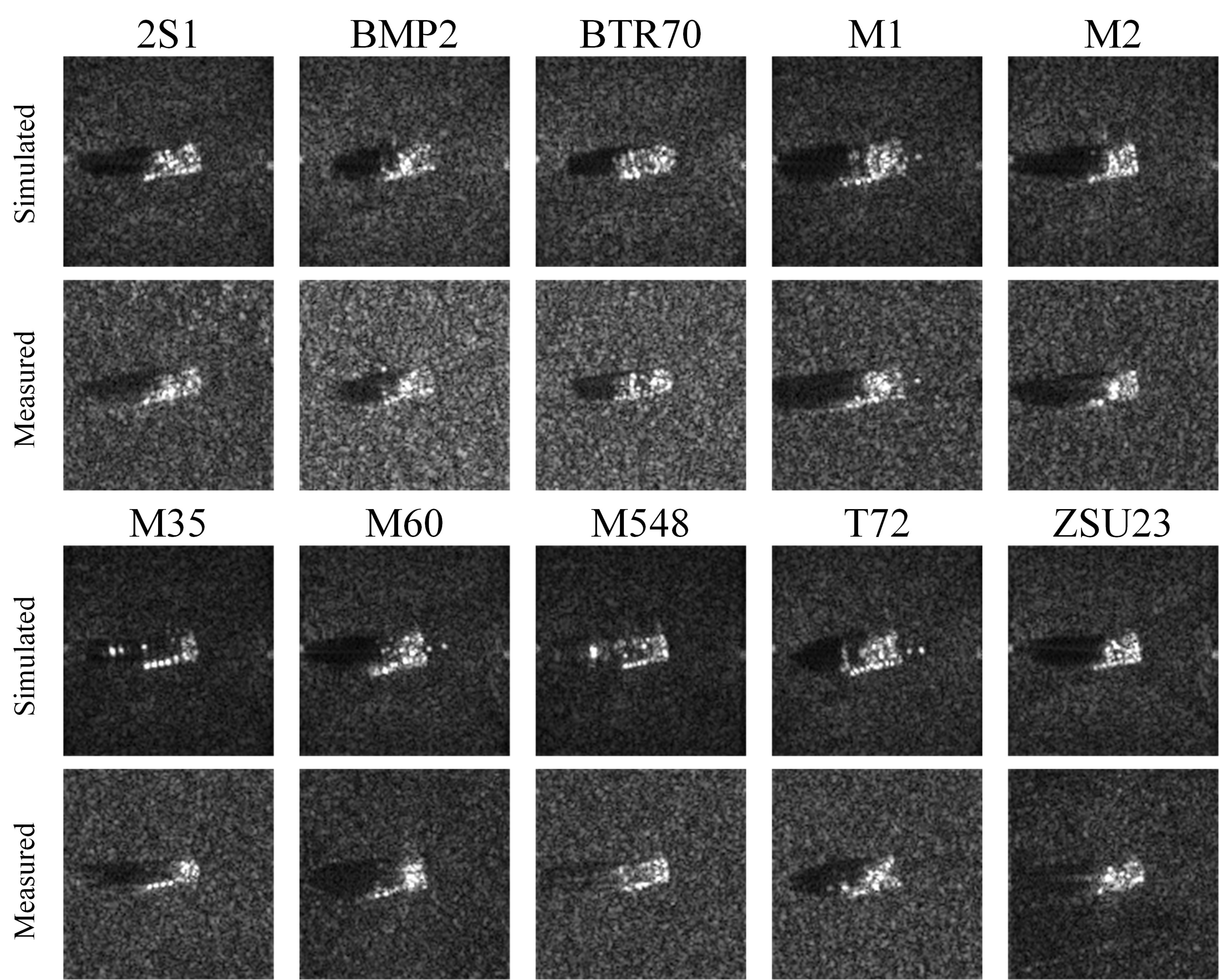}
	\caption{Examples of SAMPLE dataset image pairs (simulated-measured).}
	\label{fig_6}
\end{figure}

We evaluated the validity of the proposed approach using the publicly available SAMPLE dataset \cite{r37}, which has simulated and measured SAR image pairs of ten-class targets. The ten-class targets (2S1, BMP2, BTR70, M1, M2, M35, M548, M60, T72, and ZSU23) are identical to those of the Moving and Stationary Target Acquisition and Recognition (MSTAR) dataset. For each measured image, a corresponding simulated image was created by producing the radar echoes using EM computation. Then, using a SAR imaging algorithm, the simulated image was acquired with parameters that mimic the MSTAR dataset, such as a 0.3m distance resolution, 128 $\times$ 128 image size, HH polarization and an azimuth angle range from 10$^\circ$ to 80$^\circ$. The samples’ elevation angles ranged from 14$^\circ$ to 17$^\circ$. Table \ref{tab:table1} lists the quantity of simulated and measured SAR samples (14$^\circ$-16$^\circ$ and 17$^\circ$) for each category. Examples of the ten-class target simulated and measured image pairs are presented in Fig. \ref{fig_6}.

\begin{table}[!t]	
	\normalsize
	\renewcommand\arraystretch{1.5}
	\caption{Number of SAR samples of ten categories in the SAMPLE dataset \label{tab:table1}}
	\centering
	\setlength{\tabcolsep}{3mm}{
		\begin{tabular}{l c c c}
			\toprule[1pt]
			Target&	Simulated& 	Real(14$^\circ$-16$^\circ$)& 	Real(17$^\circ$)\\
			\hline\noalign{\smallskip}
			2S1&	174&	116&	 58\\
			BMP2&	107&	 55&	 52\\
			BTR70&	 92&	 43&	 49\\
			M1&		129&	 78&	 51\\
			M2&		128&	 75&	 53\\
			M35&	129&	 76&	 53\\
			M548&	128&	 75&	 53\\
			M60&	176&	116&	 60\\
			T72&	108&	 56&	 52\\
			ZSU23&	175&	116&	 58\\
			\hline\noalign{\smallskip}
			Total&	1345&	806&	539\\
			\bottomrule[1pt]
	\end{tabular}}
\end{table}

We established two cases for the SAMPLE dataset experiment. In Case I, which is more typical, all of the test data is measured data from 17° elevation angle, whereas the training data is made up of both simulated and measured data from 14° to 16° elevation angle. In Case II, all of the test data is made up of 14° to 16° measured data, whereas the training data is made up of simulation and 17° measured data. Compared to Case I, the training set has fewer real data while the test set contains more real data in Case II. Furthermore, the measured data in the training set of Case II has an elevation angle of just 17°, which is more homogeneous. In contrast, with an elevation angle ranging from 14° to 16°, the measured data in Case I’s test set is more varied. Consequently, it is more challenging to recognize targets in Case II.

In particular, the source domain in our experiments was considered to be the simulated SAR data with available class labels. Additionally, the measured SAR data was given very few labels and was used as the target domain. The entire target domain dataset was split into three sections in accordance with the SSDA learning paradigm: test, labeled training and unlabeled training subsets. The labeled and unlabeled training target data aid in the learning of the SSDA model in addition to the labeled source data.

The implementation specifics are as follows. In our approach, the classifier was a single FC, and a ResNet34 served as the feature extractor. The learning rates of the feature extractor and classifier were respectively 0.01 and 0.001. The decay was set as 0.0005. With a batch size of 24, the network was trained for 5000 iterations. Empirically, the $\alpha $ in \eqref{eq10}, \eqref{eq11}, \eqref{eq12} was set to 0.5, and the $\sigma $ in \eqref{eq20} was set to 0.95. The ${\beta ^2}$ in \eqref{eq21} was set to 0.5. And, the balance parameter ${\lambda _{msr}}$ in \eqref{eq23} was set to 1. The balance parameter ${\lambda _{pta}}$ and ${\lambda _{cona}}$ in \eqref{eq24} were set as 0.1 and 1, respectively.

We conducted our research using an Intel(R) Core(TM) i7-12700F CPU in a PC. The hardware and software was configured with NVIDIA GeForce RTX 2080 SUPER GPU and PyTorch 3.11.

\subsection{Comparison Experiments}
The performance of the proposed approach was evaluated by contrasting it with several existing SSDA methods, including “S+T”, “ENT”, “MME”, “CDAC” and “CDAC+SLA”. As the benchmark approach, “S+T” indicates the target classification model that is trained directly using only labeled source and target data without any domain adaptation. Another commonly comparative method is referred to “ENT”, which is based on minimum entropy principle \cite{r56}. “MME” stands for maximal minimum entropy adversarial adaptive approach \cite{r30}. “CDAC” represents the SSDA method on the basis of cross-domain adaptive clustering using the adversarial feature alignment technique \cite{r17}. “SLA” is a cutting-edge adaptive paradigm developed from the source labels denoising perspective, which is usually combined with other SSDA methods \cite{r36}. In this paper, it is jointed with “CDAC”, being denoted as “CDAC+SLA”.

We used $k$-shot to show that the target domain training subset contains $k$ labeled samples per class, and $k \in \left\{ {1,3} \right\}$. The comparison results of the 1-shot and 3-shot  in Case I are displayed in Table \ref{tab:table2}, and those in Case II are shown in Table \ref{tab:table3}. 

Table \ref{tab:table2} shows that under 1-shot condition, every approach outperformed 1-shot in Case I in terms of accuracy. It is because each class has 3 labeled target samples for the model training in the 3-shot. The “S+T” only achieved 71.40\% recognition accuracy under 1-shot condition. Apart from the baseline approach “S+T”, other methods leveraged the unlabeled target data for training. Consequently, they all performed better than the former in Case I. The “ENT” technique focused on mining unlabeled data by computing and reducing entropy on the unlabeled data. Therefore, “ENT” had a considerable performance improvement over “S+T”, achieving an average of 82.61\% recognition precision. For SSDA, “ENT” lacked contribution to the target domain’s alignment with the source domain. “MME” is a minimum-maximum entropy method that has an adversarial process and entropy maximization. Compared to “ENT”, “MME” contributed to the domain alignment and utilized the unlabeled target data more effectively. Hence, “MME” had further performance improvements, achieving an average recognition accuracy of 86.17\%. However, due to the fact that the majority of the training data are from the source domain, “MME” fails to achieve precise alignment between these two domains.  In view of this problem, “CDAC” minimizes the clustering loss in the feature extractor of the target domain to form clusters. Conversely, it maximizes the clustering loss to teach the classifiers domain-invariant features. Adversarial learning in “CDAC” facilitates both intra- and inter-domain adaptation. From a new perspective, regarding the target data, “SLA” views the source data as its noise version. It dynamically clears the noise of the source domain data, which can better mine the domain invariant information of the source domain data. Both “CDAC” and “CDAC+SLA” exhibited outstanding experimental results under 3-shot condition, obtaining 94.86\% and 96.34\% recognition accuracy respectively. Nevertheless, the aforementioned methods focus on optical images and fail to take SAR image characteristics into account. On the basis of analysing simulated and measured SAR data characteristics by the wavelet transform, our approach provided PWTDA and AIPA for SAR ATR under SSDA fashion. Among all methods, our approach delivered best performance under 1-shot condition, achieving 99.63\% precision in recognition and demonstrating superiority.

\begin{table}[!t]	
	\normalsize
	\renewcommand\arraystretch{1.5}
	\caption{Experimental results of Case I, Best results in bold. \label{tab:table2}}
	\centering
	\setlength{\tabcolsep}{3mm}{
		\begin{tabular}{c c c}
			\toprule[1pt]
			\makecell{Number of labeled\\target samples per class}& 	1&	3\\
			\hline\noalign{\smallskip}
			S+T&	71.40$\pm$2.65&	78.66$\pm$3.66\\
			ENT\textsuperscript{\cite{r56}} &	82.61$\pm$1.98&	86.55$\pm$2.42\\
			MME\textsuperscript{\cite{r30}} &	86.17$\pm$1.37&	89.41$\pm$0.48\\
			CDAC\textsuperscript{\cite{r17}} &	94.86$\pm$1.27&	98.04$\pm$0.71\\
			CDAC+SLA\textsuperscript{\cite{r36}} &	96.34$\pm$0.85&	98.17$\pm$0.87\\
			\textbf{Ours}&	\textbf{99.63$\pm$0.35}&	\textbf{99.74$\pm$0.32}\\
			\bottomrule[1pt]
	\end{tabular}}
\end{table}

\begin{table}[!t]	
	\normalsize
	\renewcommand\arraystretch{1.5}
	\caption{Experimental results of Case II, Best results in bold. \label{tab:table3}}
	\centering
	\setlength{\tabcolsep}{3mm}{
		\begin{tabular}{c c c}
			\toprule[1pt]
			\makecell{Number of labeled\\target samples per class}& 	1&	3\\
			\hline\noalign{\smallskip}
			S+T&	67.73$\pm$3.94&	78.08$\pm$3.21\\
			ENT\textsuperscript{\cite{r56}} &	82.37$\pm$3.47&	86.59$\pm$0.94\\
			MME\textsuperscript{\cite{r30}} &	85.86$\pm$3.51&	88.15$\pm$1.81\\
			CDAC\textsuperscript{\cite{r17}} &	94.70$\pm$1.76&	98.11$\pm$0.73\\
			CDAC+SLA\textsuperscript{\cite{r36}} &	96.26$\pm$1.58&	98.28$\pm$0.56\\
			\textbf{Ours}&	\textbf{98.91$\pm$0.71}&	\textbf{99.18$\pm$0.62}\\
			\bottomrule[1pt]
	\end{tabular}}
\end{table} 

As for Case II, Table \ref{tab:table3} gives the comparison results for both 1-shot and 3-shot condition. Obviously, Case II is more difficult than Case I, since the former has more test data and fewer training data with only one elevation angle. In general, the experimental results in Case II for every method are marginally inferior to those in Case I. Table \ref{tab:table3} shows that the suggested approach continued to produce the best results. Furthermore, our method can attain around 99\% recognition accuracy under both 1-shot and 3-shot, indicating its advantage in case of extremely rare labeled target data.

\begin{figure*}[htbp]
	\centering
	\includegraphics[width=0.80\textwidth]{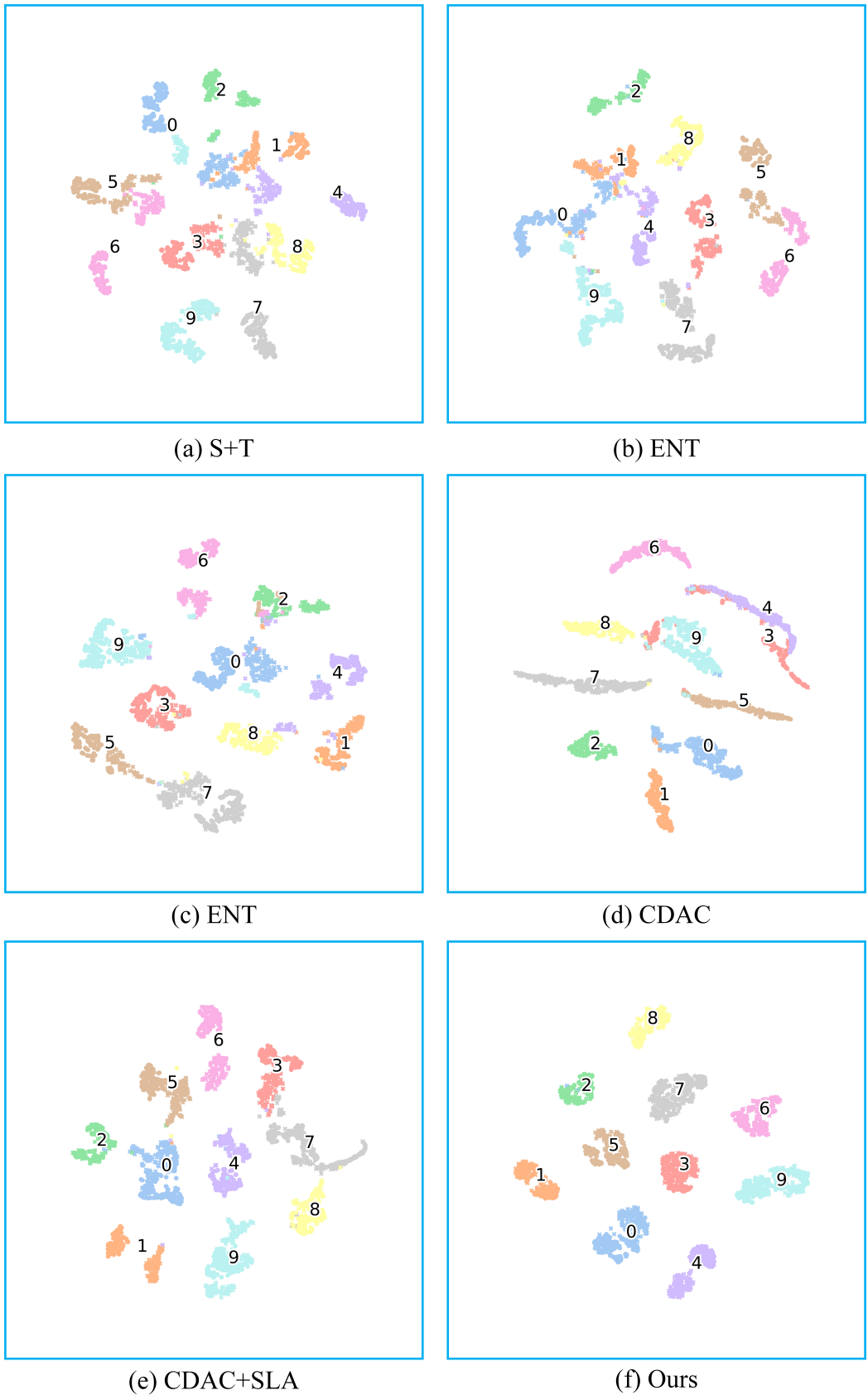}
	\caption{Visualisation of feature t-SNE for all methods in Case I for the 1-shot setting. The same colour indicates the same class and the numbers represent the centre of the categories. Simulation features are denoted by ‘$\cdot $’ and measurement features by ‘$\times $’.}
	\label{fig_7}
\end{figure*}

\begin{figure*}[!t]
	\centering
	\includegraphics[width=1.0\textwidth]{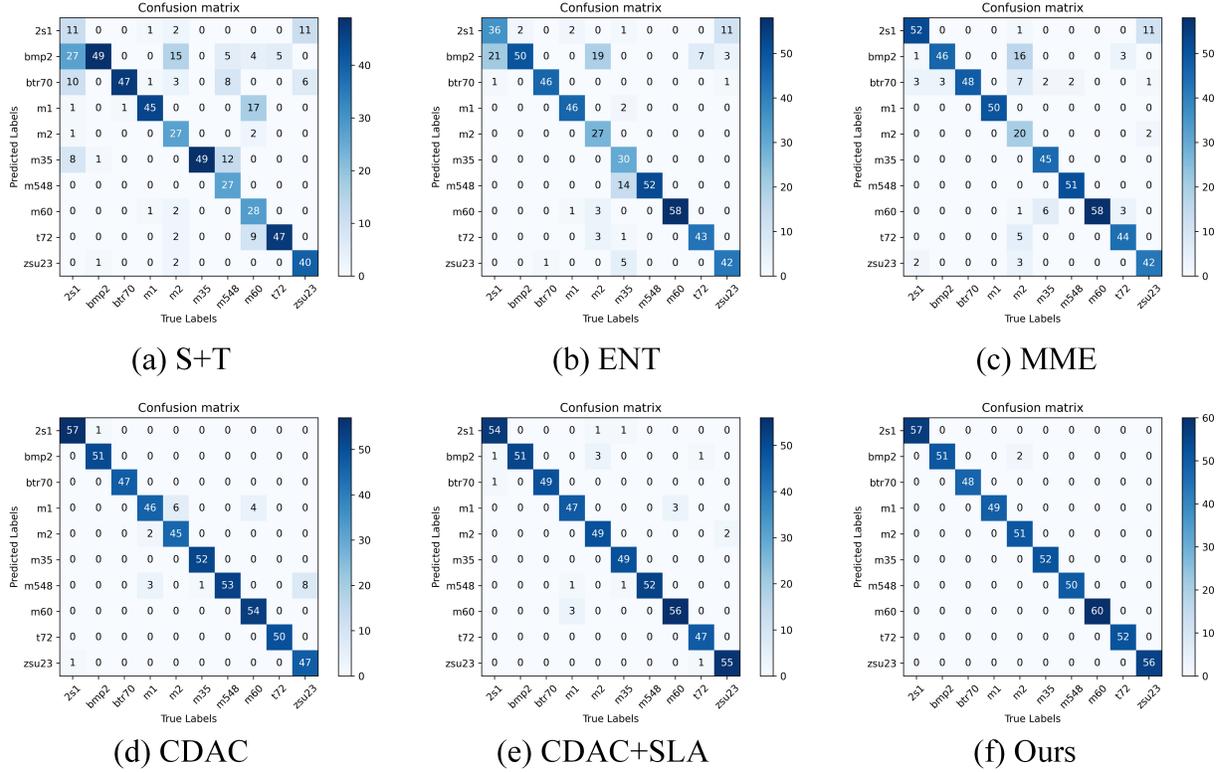}
	\caption{Confusion matrix visualisation results for all methods in Case I for the 1-shot setting. For each confusion matrix, the columns match the predicted categories, and the rows show the actual categories.}
	\label{fig_8}
\end{figure*}

Fig. \ref{fig_7} gives feature t-SNE visualization of all methods for the 1-shot setting in Case I, and Fig. \ref{fig_8} demonstrates the confusion matrix of all methods under the same setup. For “S+T”, as shown in Figs. \ref{fig_7}(a) and \ref{fig_8}(a), the model obtained the worst results on the measured domain, due to being trained only on simulated data and few measured labeled data. This is essentially because the simulated and measured domains have a significant domain shift. Compared to “S+T”, “ENT” exploited a large amount of unlabeled measured data for the whole model learning, facilitating the reduction of domain gap. More alignment operations were carried out by other methods, such as “MME”, “CDAC” and “CDAC+SLA”. Therefore, the phenomena of category confusion and domain shift were further mitigated for these three methods. Our approach greatly improves both intra-class compactness and inter-class separability, as seen in Fig. \ref{fig_7}. It means that the learned features of the target domain by our approach are more representative and discriminative than other methods. 

\subsection{Ablation Experiment}
Ablation studies were conducted in this section to assess the individual components of the method’s contribution. We conducted the ablation experiments in Case I under the 1-shot setup to confirm the effectiveness of PWTDA and AIPA. The experimental results are displayed in Table \ref{tab:table4}. It is evident that these two modules are valid for boosting the whole model performance. Table \ref{tab:table4} shows that the recognition accuracy of the model was 96.02\% when PWTDA and AIPA were unavailable. Adding PWTDA improves the accuracy to 98.95\%, demonstrating the effectiveness of the simulated and measured domain alignment at the domain-level. The accuracy rose to 98.88\% with only AIPA, demonstrating both the efficacy and ability of AIPA to achieve progressive instance-prototype alignment. The accuracy increased to 99.63\% in the presence of PWTDA and AIPA. All of results shows how successful the modules we suggested in detail.

\subsection{Hyperparameter Analysis Experiments}
\begin{figure}[!t]
	\centering
	\includegraphics[width=0.45\textwidth]{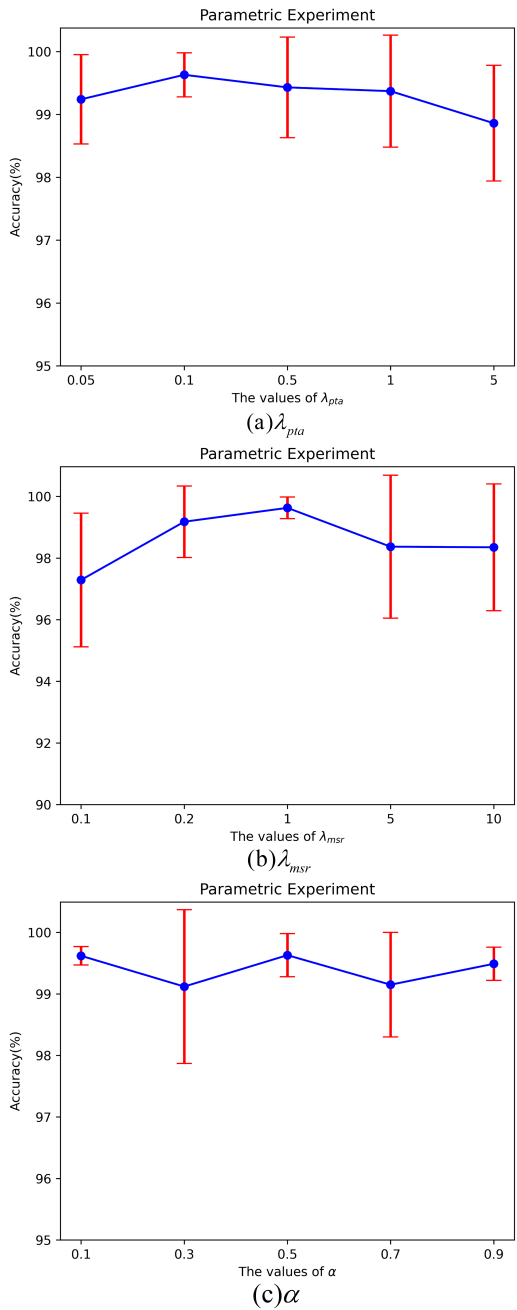}
	\caption{Parametric experiment results.}
	\label{fig_9}
\end{figure}

In this section, we examined the effects of critical hyperparameters ${\lambda _{pta}}$, ${\lambda _{msr}}$ and $\alpha $, on the performance under the 1-shot setup in Case I.

\begin{table}[!t]	
	\normalsize
	\renewcommand\arraystretch{1.5}
	\caption{The result of ablation experiment in Case I for the 1-shot setting. Best results in bold \label{tab:table4}}
	\centering
	\setlength{\tabcolsep}{4mm}{
		\begin{tabular}{c c c}
			\toprule[1pt]
			PWTDA&	AIPA& 	Accuracy\\
			\hline\noalign{\smallskip}
			--&	--&	96.02$\pm$3.86\\ 
			\checkmark&	--&	98.95$\pm$0.78\\
			--&	\checkmark&	98.88$\pm$0.69\\
			\checkmark&	\checkmark&	\textbf{99.63$\pm$0.35}\\
			\bottomrule[1pt]
	\end{tabular}}
\end{table}

We investigated the influence of ${\lambda _{pta}}$ in \eqref{eq24} with different values of ${\lambda _{pta}} \in \left\{ {0.05,0.1,0.5,1.0,10} \right\}$. In this case, the values of ${\lambda _{msr}}$ and $\alpha $ were fixed at 1 and 0.5 respectively. The analysis results of ${\lambda _{pta}}$ are presented in Fig. \ref{fig_9}(a). It can be observed that when ${\lambda _{pta}}$ varies within the range from 0.05 to 1.0, the recognition accuracy of the model remains above 99\%. When ${\lambda _{pta}}$ changes to 10, the accuracy drops below 99\%.

Fig. \ref{fig_9}(b) displays the analysis results for ${\lambda _{msr}}$ in \eqref{eq23}. The values of ${\lambda _{pta}}$ and $\alpha $ were fixed as 0.1 and 0.5 respectively. We studied the influence of ${\lambda _{msr}}$ with different values of   ${\lambda _{msr}} \in \left\{ {0.1,0.2,1,5,10} \right\}$. It is evident that ${\lambda _{msr}}$ changes around 1, having little impact on the model performance. If ${\lambda _{msr}}$ deviates largely from this range, the recognition performance is obviously reduced.

The influence of $\alpha $ in \eqref{eq10}, \eqref{eq11}, \eqref{eq12}, was investigated with different values of  $\alpha  \in \left\{ {0.1,0.3,0.5,0.7,0.9} \right\}$. In this case, the values of ${\lambda _{pta}}$ and ${\lambda _{msr}}$ were fixed as 0.1 and 1 respectively. Fig. \ref{fig_9}(c) displays the results of the analysis for $\alpha $. It is shown that the change of $\alpha $ in the PWTDA barely affects the performance. 

In brief, the settings of these hyperparameters values do not significantly affect the algorithm’s performance.

\section{Discussion}
\label{sec5}
In this research, we use synthetic data to study an SSDA approach for SAR ATR. Our approach can bridge the domain gap by performing hierarchical cross-domain alignment.

Aiming to implement domain-level alignment, we proposed the PWTDA, enabling a close relationship between the simulated and measured data on the wavelet high-frequency sub-bands. To our best knowledge, no researchers have examined cross-domain alignment in the field of SSDA for SAR ATR from the wavelet transform perspective.  However, the high-frequency components are overly homogeneous because there is a dearth of target data with labels, causing certain flaws in the model’s capacity to generalize. In order to address this issue, we incorporated high-confidence pseudo-label target data into the augmentation procedure. Hence, it is possible to enrich high-frequency components of augmented data, leading to the promoted generalization capacity. Although the PWTDA seems simple, the outcomes of the experiment show that it works well for domain-level alignment.

For achieving category-level alignment, we explored the instance-prototype alignment technique termed as AIPA. The source domain instances and their corresponding target prototypes have been aligned through the application of the instance-prototype alignment technique. Due to a lack of labeled data, target prototypes are unable to accurately reflect the true target distribution in the feature space. To solve this problem, the pseudo-label target data is included in into the computation of the target category prototype. Note that the pseudo-labels are continually updated through the model training. As the training progresses, the pseudo-label samples are gradually increased, the target prototypes closely resemble the true class distribution. Both intra-class compactness and inter-class separation can be successfully increased with this strategy.
 
The consistency alignment is considered as another categorical alignment technique. Note that it is only exploited in the measured domain. The generalizability of the model is improved using a strong data augmentation approach, and data perturbations are forced to make the model resilient through consistency learning. In particular, both single sample and multiple sample relationship consistency are taken into consideration.

Briefly, our approach delivers excellent accuracies on SAMPLE dataset. It is worth noting that out approach can achieve 99\% accuracy with only one labeled sample of the target domain, which is very meaningful for real-world SAR ATR applications. In the future, we intend to explore the application of optical images to SAR ATR under SSDA fashion.

\section{Conclusion}
\label{sec6}
In this paper, an SSDA framework is proposed for simulated data-aided SAR ATR. The framework achieves satisfied recognition accuracies for measured images with extremely scarce labeled and a large amount of unlabeled simulated data, that is important for practical SAR ATR application. The success of the proposed approach lies in collaborative multi-level alignments in a progressive manner. Thanks to the discovery that the discrepancies between simulated and measured data mainly occur at the wavelet transform high-frequency sub-bands, we develop the PWTDA to obtain the domain-level alignment. Further, the source domain samples are aligned to their corresponding target prototypes by the AIPA, not only achieving category-level alignment but also strengthening intra-class proximity and inter-class separation. Moreover, the consistency alignment is implemented in the target domain, following the consistency regularization in SSL. Note that the aforementioned multi-level alignments are collaboratively carried out through the model training process. Despite the striking recognition accuracies obtained on the SAMPLE dataset, further testing of the suggested method needs to be verified on other simulated and measured SAR datasets in future.

\bibliographystyle{IEEEtran}
\bibliography{refer}

\vspace{11pt}

\begin{IEEEbiography}[{\includegraphics[width=1in,height=1.25in,clip,keepaspectratio]{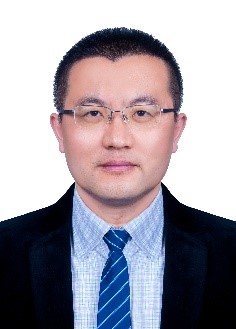}}]{Xinzheng Zhang} received the B.S. degree in automation from the Wuhan University of Technology, Wuhan, China, in 1999. He received the M.S. degree in signal and information processing, and the Ph.D. degree in communication and information system from the China Aerospace Science \& Industry Corporation, Beijing, China, in 2002 and 2009, respectively. Since 2010, he has been an Associate Professor with the School of Microelectronics and Communication Engineering, Chongqing University, Chongqing, China. His research interests include SAR remote sensing, target detection and recognition, machine learning and its application.
\end{IEEEbiography}

\begin{IEEEbiography}[{\includegraphics[width=1in,height=1.25in,clip,keepaspectratio]{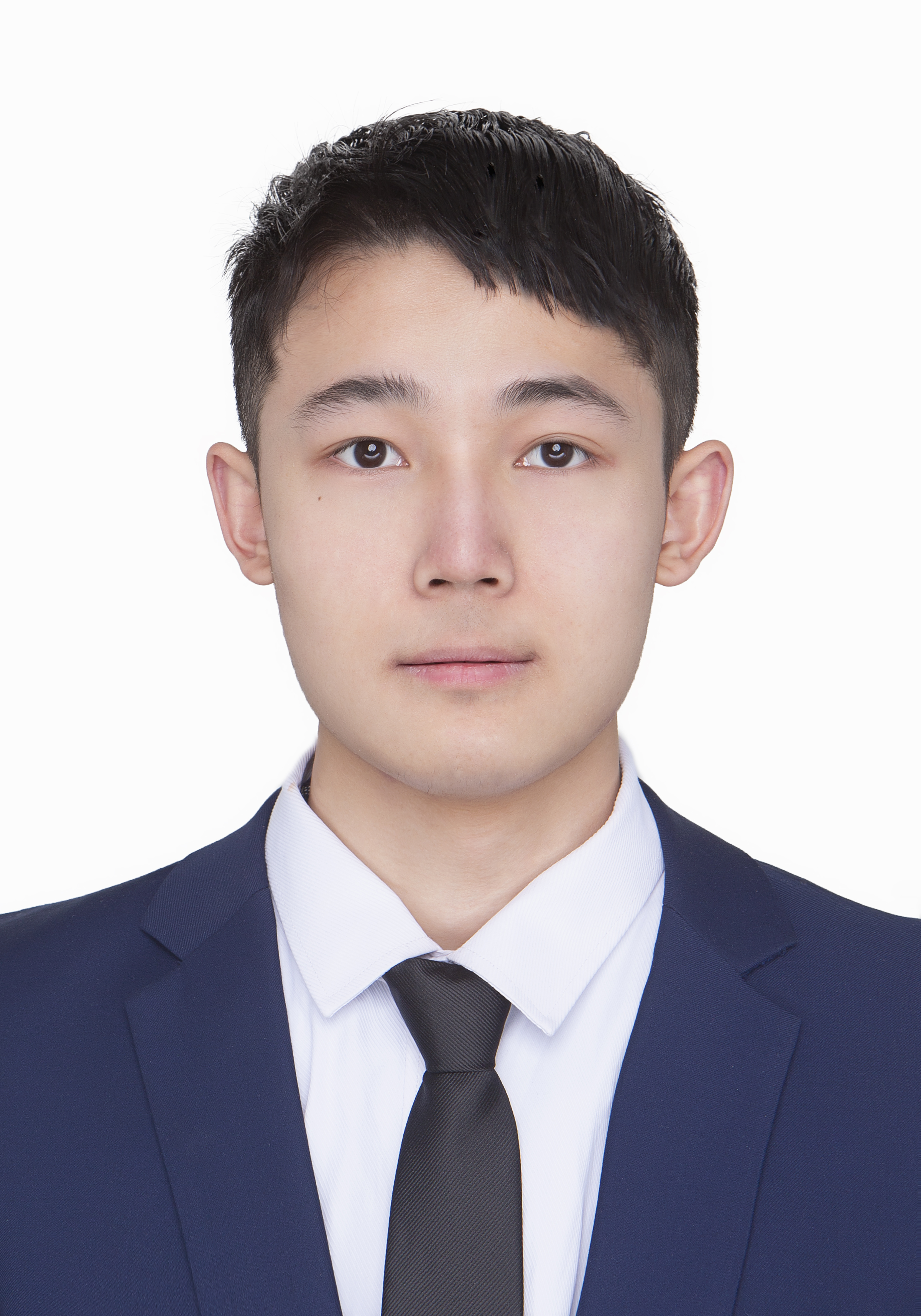}}]{Hui Zhu} received the B.S. degree from the School of Information Engineering, Zhengzhou University, Zhengzhou, China, in 2022. Currently, he is pursuing the M.S. degree with the School of Microelectronics and Communication Engineering, Chongqing University, China. His research interests include synthetic aperture radar remote sensing, deep learning, and automatic target recognition.
\end{IEEEbiography}

\begin{IEEEbiography}[{\includegraphics[width=1in,height=1.25in,clip,keepaspectratio]{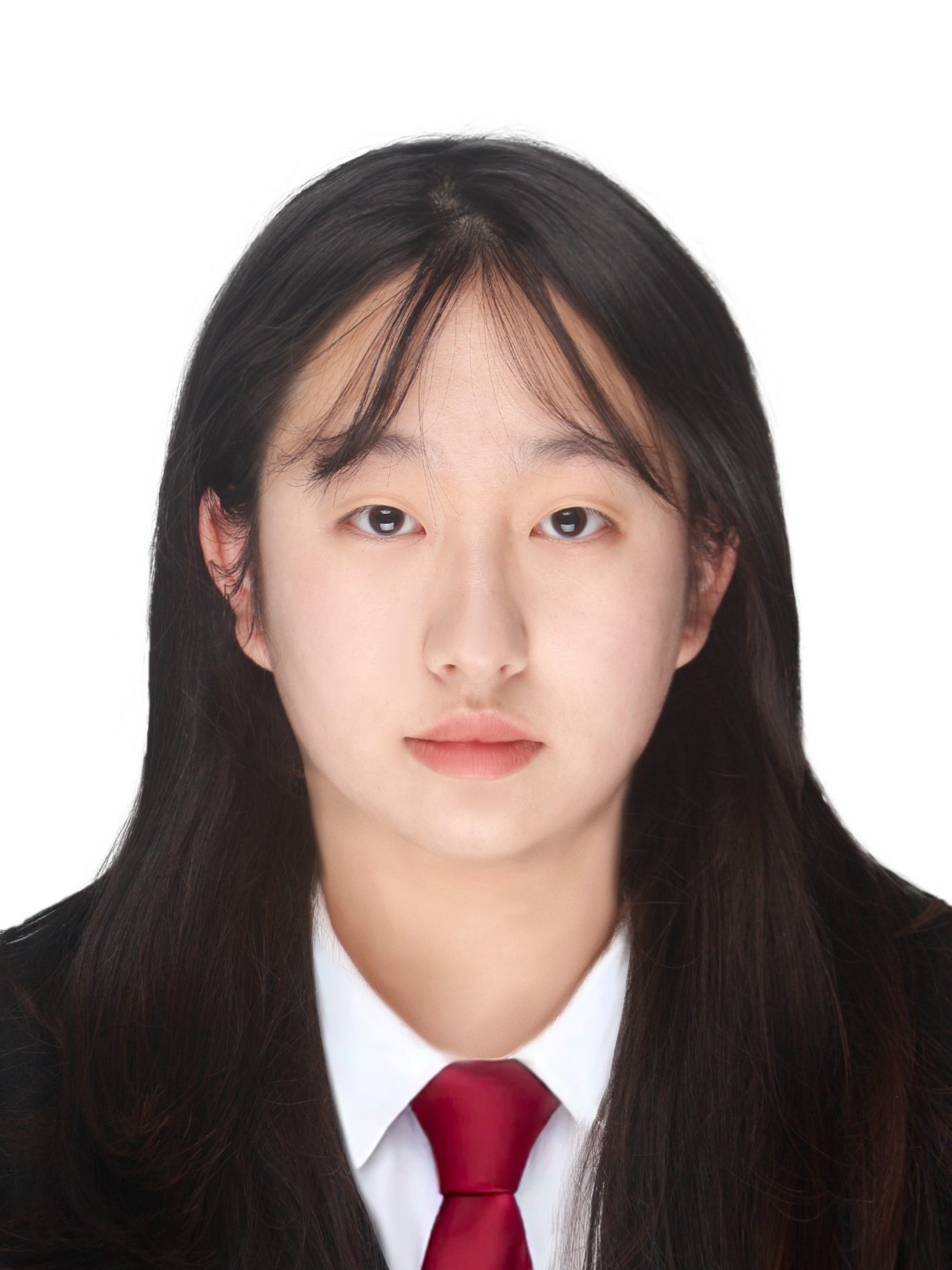}}]{Hongqian Zhuang} obtained a B.S. degree in Electronic Information Engineering from the School of Physics and Electronic Engineering at Sichuan Normal University in China in 2024, and is currently pursuing a Master's degree in Microelectronics and Communication Engineering at Chongqing University in China. Her research interests include deep learning and synthetic aperture radar remote sensing, as well as intelligent perception.
\end{IEEEbiography}

\vspace{11pt}

\vfill

\end{document}